\documentclass[letterpaper, 10 pt, conference]{ieeeconf}  

\IEEEoverridecommandlockouts                              

\usepackage{graphics} 
\usepackage{epsfig} 
\usepackage{amsmath} 
\usepackage{amssymb}  
\usepackage[ruled,linesnumbered]{algorithm2e}
\usepackage{todonotes}
\usepackage[caption=false,font=footnotesize]{subfig}
\usepackage{multirow}
\usepackage{tabularx}
\usepackage{tablefootnote}
\usepackage{footnote}
\usepackage{dblfloatfix}
\usepackage[normalem]{ulem}

\usepackage{tikz}
\usepackage{textcomp}
\usepackage{lipsum}
\usepackage[bookmarks=false]{hyperref}
\usepackage{amsmath}

\usepackage{caption}

\usepackage{graphicx}
\usepackage{cite}
\usepackage{picinpar}
\usepackage{amsmath}
\usepackage{url}
\usepackage{flushend}
\usepackage{colortbl}
\usepackage{soul}
\usepackage{multirow}
\usepackage{pifont}
\usepackage{color}
\usepackage{alltt}
\usepackage{siunitx}
\usepackage{breakurl}
\usepackage{epstopdf}
\usepackage{pbox}

\title{\LARGE \bf
Cooperative vs. Teleoperation Control of the Steady Hand Eye Robot with Adaptive Sclera Force Control: A Comparative Study
}

\author{Mojtaba Esfandiari$^{1}$, Ji Woong Kim$^{1}$, Botao Zhao$^{1}$, Golchehr Amirkhani$^{1}$, Muhammad Hadi$^{1}$, {\it Student, IEEE},\\ Peter Gehlbach$^{2}$, {\it Member, IEEE}, Russell H. Taylor$^{3}$, {\it Life Fellow, IEEE}, Iulian Iordachita$^{1}$, {\it Senior Member, IEEE} 
\thanks{*This work is supported by the U.S. National Institutes of Health under the grant numbers 2R01EB023943-04A1 and 1R01 EB025883-01A1, and partially by JHU internal funds.}
\thanks{$^{1}$ Mojtaba Esfandiari, Ji Woong Kim, Botao Zhao, Golchehr Amirkhani, Muhammad Hadi, and Iulian Iordachita are with the Department of Mechanical Engineering and also Laboratory for Computational Sensing and Robotics at the Johns Hopkins University, Baltimore, MD, 21218, USA (e-mail: {\tt\small mesfand2,jkim447,bzhao17,gamirkh1, mhadi2, iordachita@jhu.edu}).
}%
\thanks{$^{2}$ Peter Gehlbach is with the Wilmer Eye Institute, Johns Hopkins Hospital, Baltimore, MD, 21287, USA. (e-mail: {\tt\small pgelbach@jhmi.edu})
}%
\thanks{$^{3}$Russell H. Taylor is with the Department of Computer Science and also the Laboratory for Computational Sensing and Robotics at the Johns Hopkins University, Baltimore, MD, 21218, USA. ({\tt\small rht@jhu.edu})
}
}

\begin{document}
\maketitle
\thispagestyle{empty}
\pagestyle{empty}
\vspace{-1pt}
\begin{abstract}

 A surgeon's physiological hand tremor can significantly impact the outcome of delicate and precise retinal surgery, such as retinal vein cannulation (RVC) and epiretinal membrane peeling. Robot-assisted eye surgery technology provides ophthalmologists with advanced capabilities such as hand tremor cancellation, hand motion scaling, and safety constraints that enable them to perform these otherwise challenging and high-risk surgeries with high precision and safety. Steady-Hand Eye Robot (SHER) with cooperative control mode can filter out surgeon's hand tremor, yet another important safety feature, that is, minimizing the contact force between the surgical instrument and sclera surface for avoiding tissue damage cannot be met in this control mode. Also, other capabilities, such as hand motion scaling and haptic feedback, require a teleoperation control framework. In this work, for the first time, we implemented a teleoperation control mode incorporated with an adaptive sclera force control algorithm using a PHANTOM Omni haptic device and a force-sensing surgical instrument equipped with Fiber Bragg Grating (FBG) sensors attached to the SHER 2.1 end-effector. This adaptive sclera force control algorithm allows the robot to dynamically minimize the tool-sclera contact force. Moreover, for the first time, we compared the performance of the proposed adaptive teleoperation mode with the cooperative mode by conducting a vessel-following experiment inside an eye phantom under a microscope.           

\end{abstract}

\section{INTRODUCTION} \label{Introduction}
Retinal vein occlusion (RVO) is the second most prevalent retinal vascular disease. It occurs due to a retinal vein occlusion, leading to severe vision loss. In  2015, an overall 28.06 million people worldwide (0.77$\%$ of people aged
30-89 years) were affected by RVO, branch RVO (BRVO) (23.38 million), and central RVO (CRVO) (4.67 million) \cite{song2019global}. There is no standard method for directly treating RVO with surgery, as it requires consistently and safely performing retinal vein cannulation (RVC). Due in part to human hand tremor limits, this is not broadly possible without robotics-assisted surgical systems. The diameter of the retinal veins is on the order of $150 \pm 15 \mu$m \cite{goldenberg2013diameters}, while, the root mean square (RMS) value of hand tremor of an ophthalmic surgeon is reported as $182 \mu$m \cite{riviere2000study}, which is comparable to the vein diameter. Therefore, performing freehand RVC is both challenging and potentially unsafe using free-hand methods. 

To address this issue, researchers developed several surgical robotic systems such as the Steady Hand Eye Robot (SHER) \cite{uneri2010new}, a hybrid
parallel-serial micro-manipulator \cite{nasseri2013introduction}, Preceyes \cite{van2009design}, RVRMS \cite{jingjing2014design}, and others \cite{gijbels2014experimental, ergeneman2011wireless, xiao2019classifications}, trying to mitigate (filter) physiological hand tremor and also to provide surgeons with a higher level of skill and positioning accuracy. 
 
\begin{figure}[t!]
    \centering
    \includegraphics[scale=0.29]{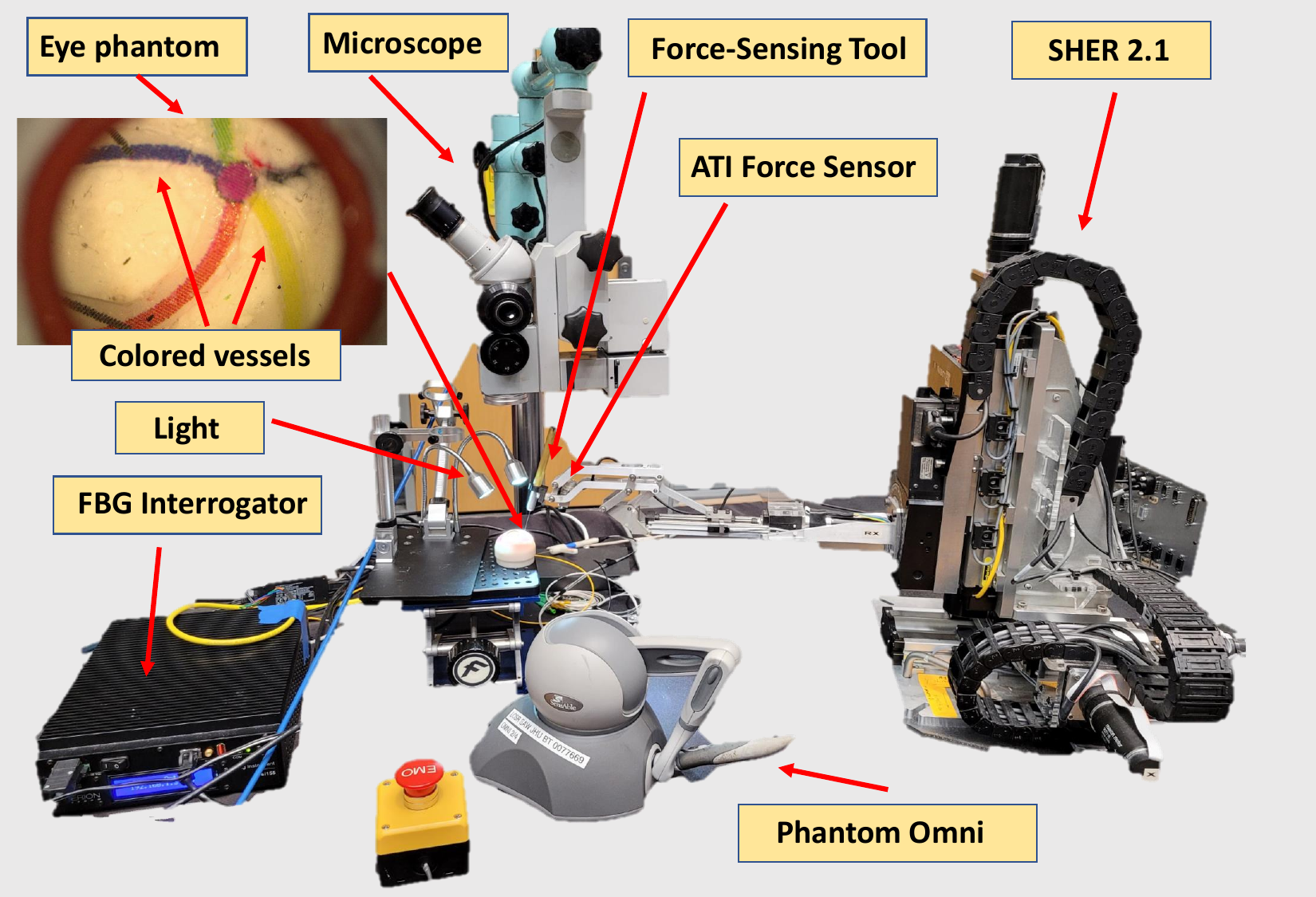}
    \caption{Vessel following experiment in an eye phantom with two different control modes. The experiment setup includes the SHER 2.1, PHANTOM Omni, a surgical microscope, an FBG interrogator, a force-sensing surgical instrument, an eye phantom, and a light source.}
    \label{fig:systems_setup_unilat_tele_coop}
\end{figure}

 The SHER is a 5-degree-of-freedom (DoF) robot manipulator that allows surgeons to cooperatively manipulate a surgical instrument attached to its end-effector towards a desired target (Fig. \ref{fig:systems_setup_unilat_tele_coop}). Due to the high structural rigidity of the robot, it removes surgeon hand tremors, but the small interaction force present between the surgical instrument and the sclera entry port is diminished during cooperative (admittance) control mode. Thus the contact force exerted by the surgeon on the sclera surface can exceed a safe threshold of 120 mN \cite{ebrahimi2018real}.

 To minimize the sclera force exerted on the needle, Ebrahimi et al. developed adaptive force control algorithms, including adaptive norm control (ANC) and adaptive component control (ACC), using a force-sensing instrument equipped with fiber Bragg grating (FBG) to maintain the scleral forces below a safe threshold \cite{ebrahimi2019adaptive, ebrahimi2021adaptive}. FBG sensors can be used for shape sensing \cite{amirkhani2023design} or force sensing \cite{he2014multi} in flexible needles or continuum robots. These concepts help to dynamically enforce the remote center of motion (RCM) constraint on the eyeball and reduce the risk of tearing the sclera during surgery. However, this adaptive force control algorithm only works in the cooperative control mode. Therefore the surgeon must still cooperate directly with the robot handle. Incorporating such force control algorithms in a teleoperation modality may provide surgeons with advanced capabilities and improve patient safety \cite{osa2015hybrid, feizi2021robotics}. 
We calibrated an FBG-based force-sensing instrument based on a method provided by \cite{he2014multi} to estimate the tip force, the sclera force, and the insertion depth by measuring the wavelength shifts of the FBG sensors.
 The contributions of this paper are as follows: 

 \begin{itemize}
     \item We developed for the first time, to the best of our knowledge, a teleoperation control mode with an adaptive sclera force control algorithm for retinal surgery applications and implemented this adaptive teleoperation algorithm on the SHER 2.1 system. This teleoperation control mode has several capabilities including but not limited to 1- indirect manipulation of the robot using a haptic interface, 2- scaling down the surgeon's hand motion to increase positioning accuracy, 3- reposition of hand frames to increase flexibility and comfort, 4-  dynamically maintaining the RCM constraint and minimizing the tool-sclera interaction forces.
     \item We compared, for the first time in robot-assisted retinal surgery applications, the performance of the proposed teleoperation mode incorporated with adaptive sclera force control with a cooperative mode equipped with the same adaptive force control functionality. We conducted a vessel-following experiment inside a plastic eye phantom under a surgical microscope to compare the performance of the proposed adaptive teleoperation mode with the cooperative mode.  
 \end{itemize}
    Results are analyzed to study the pros and cons of the different control modes. 
 
 The rest of this paper is organized as follows. Robot kinematics and control algorithms are formulated in Section \ref{Kinematics}. The experimental setup and experiment procedure are explained in Section \ref{Experimental_setup}. The experimental results are analyzed in Section \ref{sec_experimental_results}. Section \ref{discussion} draws a discussion on the results and conclusion of the paper.

\vspace{-8pt}
\section{Robot Kinematics} \label{Kinematics}

\begin{figure*}
    \centering
    \includegraphics[scale=0.38]{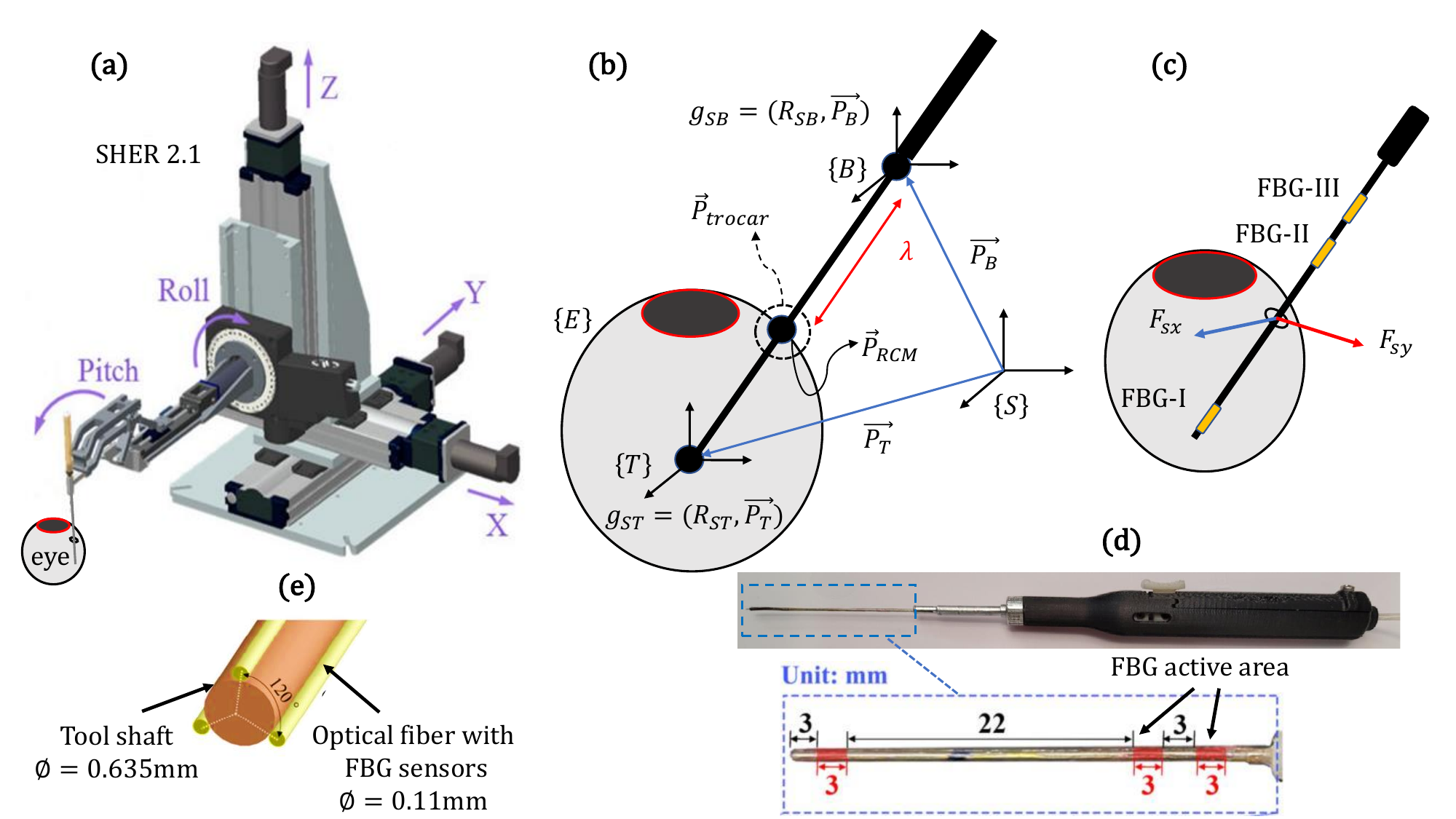}
    \caption{(a) The Steady-Hand Eye Robot, (b) RCM constraint, tooltip coordinate frame $\{T\}$, body coordinate frame $\{B\}$, and the spatial coordinate frame $\{S\}$, (c) fiber Bragg gratings sensors attached across the needle, (d) force-sensing instrument equipped with FBG sensors, and (e) cross-section of the force-sensing needle equipped with three channels of sensors  \cite{he2020automatic}.}
    \label{fig:RCM_constraint}
\end{figure*}

The Steady-Hand Eye Robot is a 5-DoF robot manipulator, including three translational and two rotational degrees of freedom. A surgical instrument, e.g., a needle, can be attached to the robot end-effector. There is no roll motion about the needle axis, however.        
The SHER motion is represented using two coordinate frames: the spatial coordinate $\{S\}$, a fixed frame in space located in the robot base, and the body coordinate $\{B\}$, which is rigidly attached to the robot handle and moves with it. The tip coordinate $\{T\}$ attached to the needle tip has the same orientation as coordinate $\{B\}$ but with a constant shift along the negative $Z$ direction of $\{B\}$ (Fig. \ref{fig:RCM_constraint}).


\vspace{-10pt}
\subsection{Kinematics of the SHER}
In this section, the forward kinematics of the SHER are developed using the homogeneous representation $ g_{SB}: \Theta \rightarrow SE(3)$ which maps the robot joints variables $\Theta \in \mathbb{R}^5$ to the robot end-effector coordinate, where $g_{SB} \in \mathbb{R}^{4\times 4}$ represents the configuration of body coordinate $B$ relative to the spatial coordinate $S$ and is a matrix in the special Euclidean group $SE(3)$. To compute the forward kinematics, we will use the product of the exponential formula \cite{murray2017mathematical} as follows 
\begin{equation} \label{eqtwist1}
g_{SB}(\Theta) = e^{\hat{\xi}_1\theta_1}...e^{\hat{\xi}_5\theta_5}g_{SB}(0)
\end{equation}
in which $\hat{\xi}_i \in se(3)$ denotes the $i^{th}$ twist, $\theta_i$ is the $i^{th}$ element of the joints variable vector $\Theta$, and $g_{SB}(0) \in \mathbb{R}^{4\times 4}$ represents the configuration of $B$ relative to $S$ when the robot is at its home position $(\Theta=\textbf{0})$.

The twist $\hat{\xi}_i$ for the first three prismatic joints ($i=1,2,3$) and for the last two rotational joints ($i=4,5$) is defined as follows \cite{murray2017mathematical} 
\begin{equation} \label{eqtwist2}
  \hat{\xi}_i = \begin{cases} \quad \begin{bmatrix}
0_{3 \times 3}       & v_i \\
0_{3 \times 1}       & 0  \\
\end{bmatrix}, \quad \quad \quad \quad \textrm{for} \quad  i = 1,2,3 \vspace{0.2cm} \\ 
\quad \begin{bmatrix}
\hat{\omega}_i       & -\omega_i \times q_i \\
0_{3 \times 1}      & 0  \\
\end{bmatrix}, \quad \textrm{for} \quad i = 4,5
\end{cases}
\end{equation}
in which $v_i$ denotes the unit vector along the positive direction of the $i^{th}$ prismatic joint and $\omega_i$ is the unit vector along the positive direction of the rotation axis (counterclockwise) of the $i^{th}$ rotational joint, both expressed in the $S$ coordinate when the robot is at home position. $q_i$ is an arbitrary point on the corresponding rotation axis. The twist coordinate $\xi_i=(v_i,\omega_i)\in \mathbb{R}^{6}$ corresponding to the twist $\hat{\xi}_i$ can be computed as $\xi_i= (\hat{\xi}_i)^{\vee}$ where $\vee$ is the vee operator\cite{murray2017mathematical}.    

Using \eqref{eqtwist1} and \eqref{eqtwist2}, the robot joint velocity vector $\dot{\Theta}\in\mathbb{R}^5$ can be mapped to the robot end-effector velocity vector $V_{SB}^b \in \mathbb{R}^6$ (the superscript $b$ means that velocity of the body coordinate $B$ relative to the spatial coordinate $S$ is expressed in the body coordinate $\{B\}$) as follows 
\begin{equation} \label{eqtwist4}
V_{SB}^b = J_{SB}(\Theta)\Dot{\Theta}
\end{equation}
where the robot Jacobian $J_{SB}\in \mathbb{R}^{6 \times 5}$ is computed as  
\begin{align}\label{eqtwist5}
J_{SB}(\Theta) &=
\begin{bmatrix}
\xi_1^\dagger & ... & \xi_5^\dagger 
\end{bmatrix},\\
\xi_i^\dagger &= Ad^{-1}_{e^{\hat{\xi}_i\theta_i}...e^{\hat{\xi}_5\theta_5}g_{SB}(0)}\xi_i \nonumber
\end{align}
and $Ad_{g}:\mathbb{R}^6 \rightarrow \mathbb{R}^6$ is the adjoint transformation associated with a rigid body transformation $g=(p,R)\in SE(3)$ defined as \cite{murray2017mathematical}
\begin{equation}
Ad_{g} = \begin{bmatrix}
R & \hat{p}R \\ 
0 & R 
\end{bmatrix}_{6\times 6}.
\label{eq: adjoint_transformation}
\end{equation}

\subsection{Cooperative Control of the SHER}

The cooperative control mode of the SHER is based on an admittance control method that lets the surgeons intuitively maneuver the robot end-effector toward a desired target. To do this, the user hand force/torque exerted on the robot handle $F_h^b\in \mathbb{R}^6$ is measured by a six DoF force sensor (Nano17, ATI Industrial Automation, NC, USA), Fig. \ref{fig:systems_setup_unilat_tele_coop}, and used in the admittance control law to generate a desired end-effector velocity in the robot body coordinate $V_d^b\in \mathbb{R}^6$ as follows 
\begin{equation} \label{eqimpedance}
V_d^b = \mathbb{K} F_h^b 
\end{equation}
$\mathbb{K}\in \mathbb{R}^{6 \times 6}$ is a diagonal matrix with constant diagonal elements as the admittance gains. Having $V_d^b$, the desired joints' angular velocity vector $\dot{\Theta}_d\in \mathbb{R}^5$ can then be calculated as 
\begin{equation} \label{eqjointangle}
\Dot{\Theta}_d = J_{SB}^\dagger(\Theta)V_d^b 
\end{equation}
and be passed to the SHER's low-level controller to generate the desired motion. $J_{SB}^\dagger$ is the Jacobian pseudo inverse \cite{chiacchio1991closed} and is calculated as  
\begin{equation}
J^{\dagger}_{SB} = J^T_{SB}(J_{SB}J^T_{SB})^{-1}.
\label{eq: Jscobian_pseudo_inverse}    
\end{equation}






\subsection{Adaptive Force Control of the SHER}
In this control algorithm, if each of the measured sclera force components, $F_{sx}$ and $F_{sy}$, exceed the specified safety threshold, $T_s$, the robot is controlled toward a safe sclera force trajectory \cite{patel2022force}. By estimating the sclera tissue stiffness, $\alpha_{i}$ $(i=x$ or $y)$, the desired sclera force trajectory, $f_{dx}$ and $f_{dy}$, can be obtained using \eqref{x-component of the desired force trajectory} and \eqref{y-component of the desired force trajectory}, respectively:
\begin{equation}
\label{x-component of the desired force trajectory}
    \begin{aligned}
    f_{dx}=\frac{T_{s}sign(F_{sx})}{2}(e^{(t-t_{x})}+1),
    \end{aligned}
\end{equation}
\begin{equation}
\label{y-component of the desired force trajectory}
    \begin{aligned}
    f_{dy}=\frac{T_{s}sign(F_{sy})}{2}(e^{(t-t_{y})}+1),
    \end{aligned}
\end{equation}
where $t_{i}$ is the time when $F_{si}$ exceeds the safety threshold.

The desired linear velocity of the end-effector along the $x$ and $y$ directions in the body frame $\{B\}$ can be determined using \eqref{AFC desired linear velocity} and \eqref{AFC tissue stiffness}:
\begin{equation}
\label{AFC desired linear velocity}
    \begin{aligned}
    \dot{X}^{b}_{des_{i}}(t)=\alpha_{i}\dot{f}_{di}(t)-K_{fi}\Delta{f}_{i}(t),\quad i=\{x,y\}, 
    \end{aligned}
\end{equation}
\begin{equation}
\label{AFC tissue stiffness}
    \begin{aligned}
    \dot{\alpha}_{i}=-\Gamma_{i}\dot{f}_{di}(t)\Delta{f}_{i}(t),\quad i=\{x,y\}
    \end{aligned}
\end{equation}
where $\Delta{f}_{i}=F_{si}-f_{di}$ is the sclera force error, and $\Gamma_{i}$ and $K_{fi}$ are constant gains for adaptation law and the force tracking error, respectively. $\alpha_{i}$ is the estimated tissue compliance, and $\dot{f}_{di}$ is the derivative of $f_{di}$ which is derived using \eqref{x-component of the desired force trajectory} and \eqref{y-component of the desired force trajectory}. The desired end-effector velocity components in the $x$ and $y$ directions, using \eqref{AFC desired linear velocity} and \eqref{AFC tissue stiffness}, are considered the desired robot end-effector velocity. This adaptive sclera force control algorithm applies to both cooperative and teleoperation control modalities. Therefore, the other elements of the robot end-effector velocity will be calculated by abiding by the user's interaction forces in the cooperative control mode, and by the PHANTOM Omni's end-effector velocity (mapped to the SHER body coordinate $\{B\}$) in the teleoperation control mode.
\begin{figure*}[!t]
    \centering
    \centerline{\includegraphics[width= 1.6 \columnwidth]{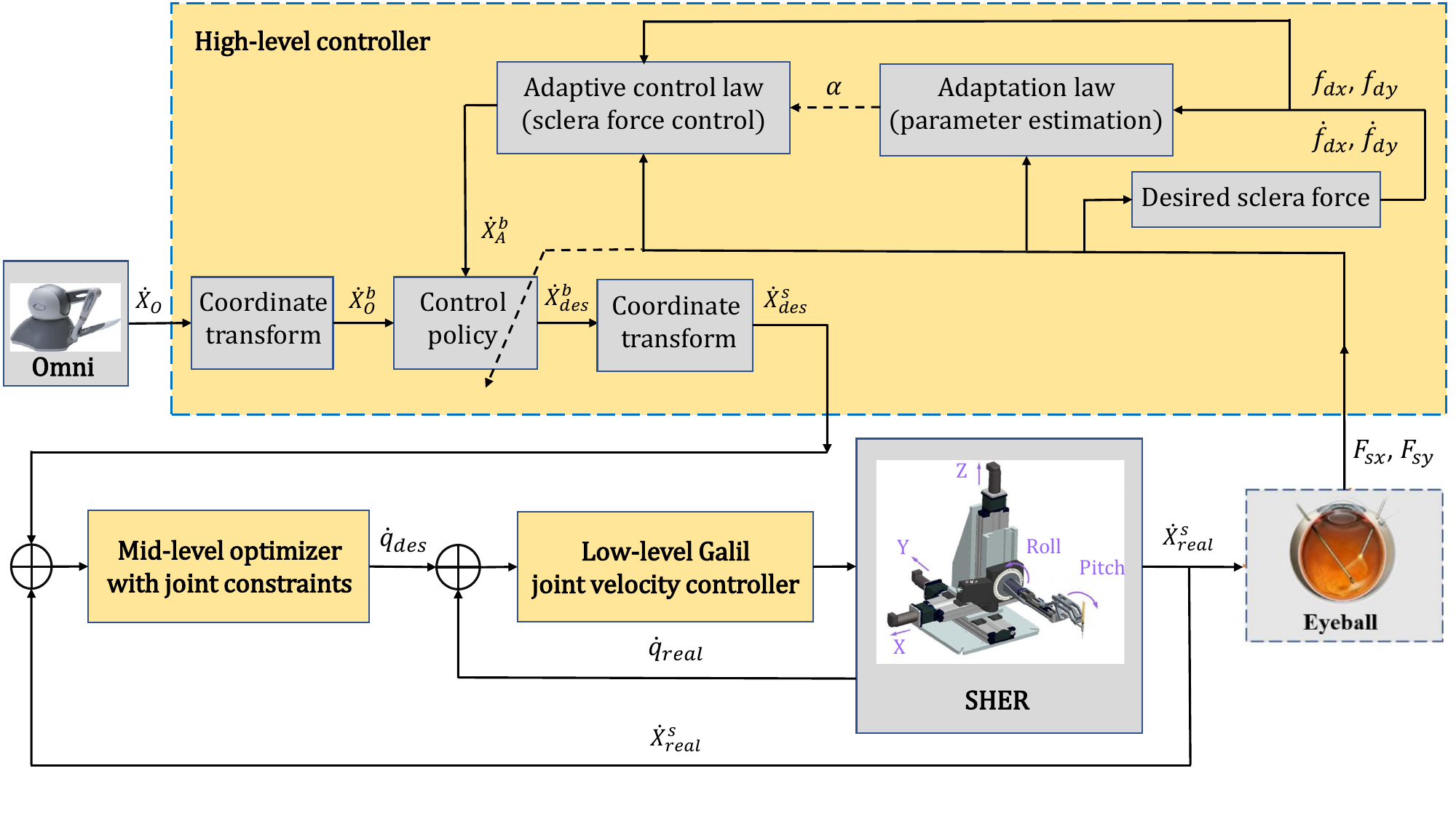}}
    \caption{Block diagram of the proposed teleoperation control architecture with adaptive sclera force control algorithm.  }
    \label{fig_Unilateral_adapticve_Tele_control_structure}
\end{figure*}

Fig. \ref{fig_Unilateral_adapticve_Tele_control_structure} shows a schematic of the proposed teleoperation control mode featuring an adaptive sclera force control algorithm. This architecture comprises three main components: a high-level controller, a mid-level optimizer with joint constraints, and a low-level joint velocity controller (Galil). 
The high-level controller consists of two key elements: an adaptive sclera force controller that takes input from the sclera force components $F_{sx}$ and $F_{sy}$ measured in the body coordinate system ${B}$ by the force-sensing instrument, which, generates the desired adaptive end-effector velocity in the SHER body coordinate as $\dot{X}_A^b$ ($=\dot{X}^{b}_{des}$ in \eqref{AFC desired linear velocity}), along with a kinematic controller that maps the end-effector velocity of the Omni, denoted as $\dot{X}_o$, to the SHER body coordinate as $\dot{X}_o^b$.
The control policy chooses between $\dot{X}_A^b$ and $\dot{X}_o^b$ depending on the magnitude of the sclera force. It then generates the desired end-effector velocity in both the SHER body coordinate ($\dot{X}_{des}^b$) and the SHER spatial coordinate ($\dot{X}_{des}^s$). The mid-level optimizer calculates the optimal desired joint rates, denoted as $\dot{q}_{des}$, for the SHER while considering its joint limit constraints. Lastly, the low-level Galil joint velocity controller produces the necessary control commands to achieve the desired joint velocities. Other signals include $\dot{q}_{real}$ (absolute joint velocities of SHER), $\dot{X}_{real}^s$ (actual end-effector velocity in the spatial coordinate of SHER), and $f_{dx}$ and $f_{dy}$ (desired sclera force components in the body coordinate ${B}$).

\vspace{-1pt}
\section{Experiments} \label{Experimental_setup}

\subsection{Experimental Setup}
 Due to the recommendation of our clinical lead, we decided to conduct a "vessel-following" experiment \cite{he2019preliminary}, which is a typical task in retinal surgery, for validation experiments and to compare the performance of four different control modes: cooperative mode (without enforcing adaptive sclera force control), adaptive cooperative mode (with enforcing adaptive sclera force control), teleoperation mode, and adaptive teleoperation mode. 
The setup is prepared to measure and record the SHER's kinematic information, the interaction force between the user's hand and the SHER handle, and the tool-sclera interaction force.  
The experimental setup includes the SHER 2.1, a PHANTOM Omni robot (SensAble Technologies Inc., MA, USA), an ATI force sensor (Nano17, ATI Industrial Automation, NC, USA), a force-sensing tool with nine FBG sensors (Technica Optical Components, China), an FBG interrogator (HYPERION si155, Luna), an eye phantom, a surgical microscope (Zeiss, OPMI), and an LED dual-light illuminator as shown in Fig.\ref{fig:systems_setup_unilat_tele_coop}. 
All devices were connected to a main computer using a TCP-IP connection.
The PHANTOM Omni is used as a joystick controller to teleoperate the SHER 2.1. and a 6-DoF ATI Nano 17 force sensor is embedded with the SHER's end-effector to measure the hand-tool interaction forces and torques from all three Cartesian coordinates.
An FBG force-sensing tool \cite{he2019dual} is used for measuring the sclera force applied to the eye phantom by the surgical instrument. The tool shaft has three FBG fibers arranged at 120-degree intervals around it. A total number of nine FBG sensors are contained in the three fibers as shown in Fig. \ref{fig:RCM_constraint}e-d. The FBG interrogator receives those optical signals from the FBG sensors and converts them into force readings.
The eye phantom is an anatomical eye model that imitates critical features of a natural human eye. This experiment is performed under an LED dual-light illuminator to get a high-contrast view from the microscope during the manipulation.
\vspace{-1pt}
\subsection{Experimental Procedure}
To perform the vessel-following experiment, the user is asked to follow a set of specific procedures \cite{he2019preliminary} and track a colored trajectory on the retinal surface of an eye phantom simulating retinal veins, for each of the four different control modes: cooperative, adaptive cooperative, teleoperation, and adaptive teleoperation modes. The vessels' pattern includes four different colors: red (R), green (G), blue (B), and yellow (Y). The experiment includes 5 sets of tests, and each includes 5 trials, a total of 25 trials for each control mode. Each trial is conducted by tracking in random order, the RGBY colors. All tests are performed under a microscope using an illuminator for best visualization. During the cooperative control mode, the user directly manipulates the SHER's tool handle, whereas, during the teleoperation mode, the user indirectly manipulates the SHER using the PHANTOM Omni. In all control modes, the user manipulates the force-sensing tool with the dominant hand and holds a secondary tool with the non-dominant hand to re-orient the eye phantom under the microscope. Of note, it is typical to perform retinal surgery bimanually.
All the kinematic and force data measured from the robot and the force-sensing tool are collected in a .csv file and analyzed in MATLAB.   


\begin{figure}[!b]
    \centering
    \includegraphics[scale=0.38]{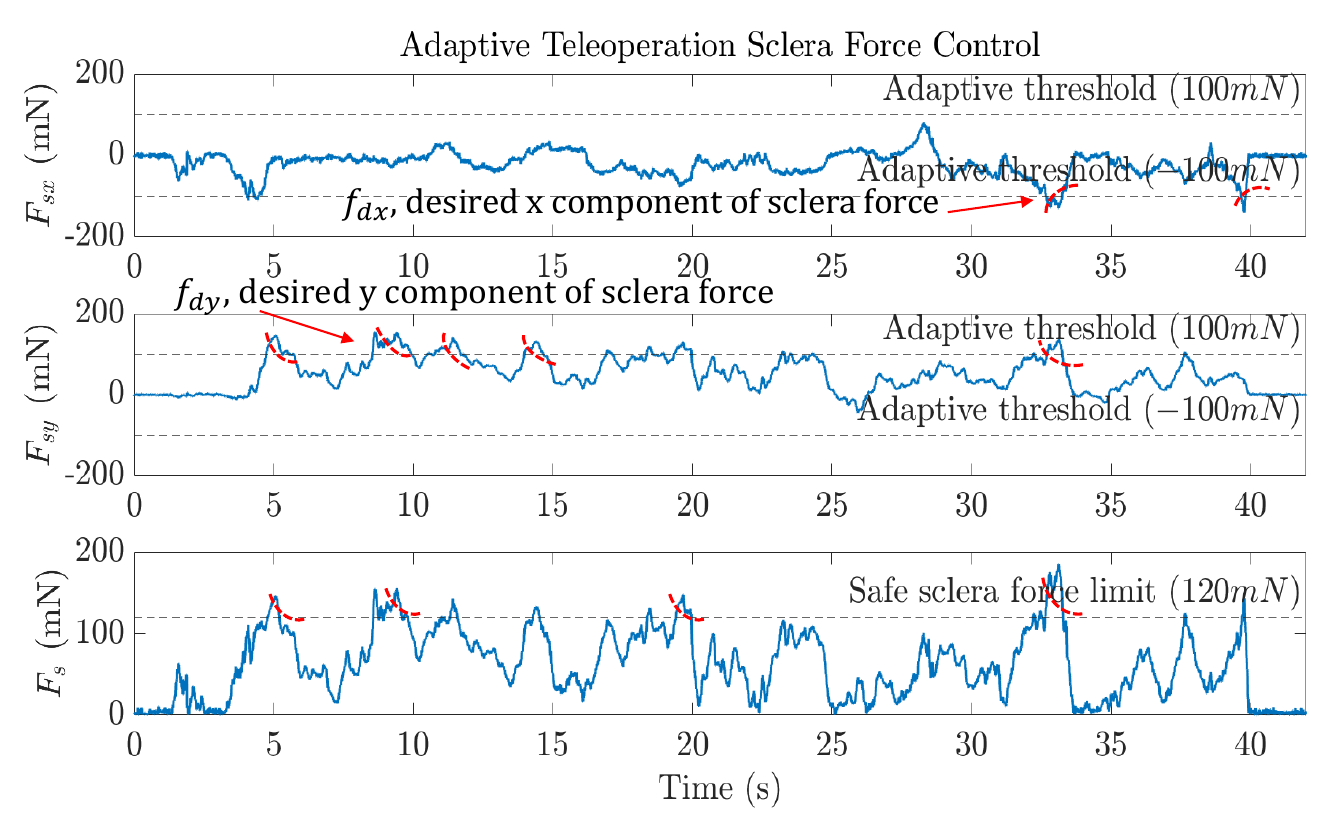}
    \caption{Example of sclera force components $F_{sx}$ and $F_{sy}$ and sclera force norm $F_s$ for adaptive teleoperation control mode. The adaptive sclera force controller is activated if $|F_{sx}|$ or $|F_{sy}|$ reaches a threshold value of 100 mN. The goal of the adaptive force controller is to maintain the sclera norm force $F_s$ below a safe threshold (120 mN) \cite{ebrahimi2018real}.}
    \label{fig_Sclera_force_adaptive_teleoperation_unilat}
\end{figure}


\vspace{-1pt}


\vspace{-1pt}
\section{Experimental Results and Discussion} \label{sec_experimental_results}
To the best of our knowledge, this is the first time a teleoperation control mode incorporated with an adaptive sclera force control algorithm has been implemented on a surgical robotic system, the SHER 2.1, for retinal surgery applications.
The primary comparison between the performance of the four control modes is studied by considering several key factors that are of notable importance in robot-assisted eye surgery, such as the tool-sclera interaction force, which is essential for the safety of the patient, and the human-robot interaction force (represented as the handle force) and the task completion time, which relates to the surgeons' comfort and intuition level. Also, a higher completion time generally results in more operational costs in an actual operating room.  
\begin{figure}[!t]
    \centering
    \includegraphics[scale=0.45]{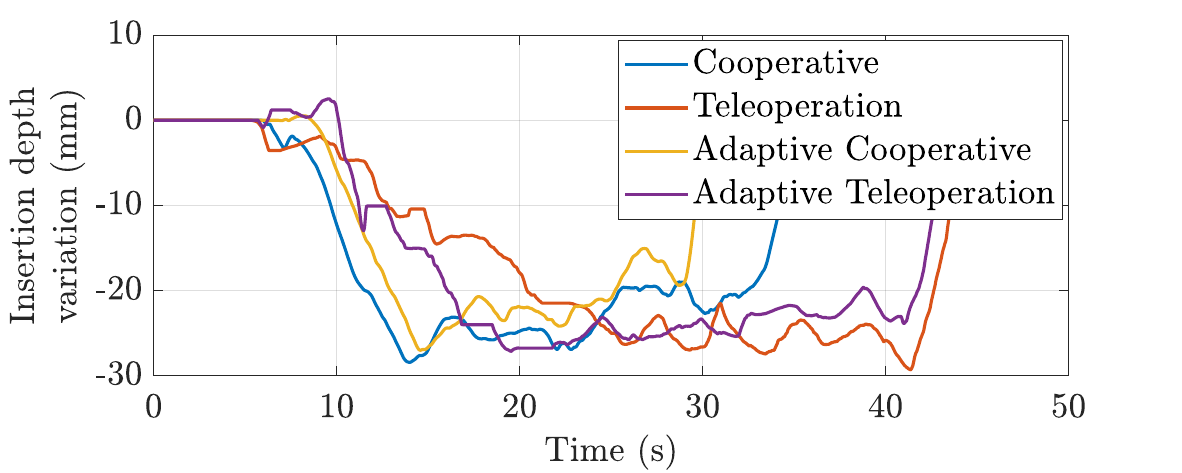}
    \caption{Example of tooltip insertion depth during the vessel-following experiment. }
    \label{fig_Insertion_depth_coop_tele_unilateral}
\end{figure}


The user is asked to hold the needle tip as close as possible to the bottom of the eye phantom and above the colored veins without touching them. Fig. \ref{fig_Insertion_depth_coop_tele_unilateral} shows an example of the tooltip insertion depth. The average value of the maximum insertion depth for all control modes is greater than 25 mm demonstrating that the user performed the vessel-following with consistent insertion depth and trajectory tracking for all four control modes. Of note, the eye phantom used in this experience has a 32 mm diameter while the human eye diameter is about 24 mm.
\begin{table}[t!]
\centering
    \caption{Mean and maximum sclera force, handle force/torque, and the completion time for the four control modes during the vessel-following tests. The values inside the parenthesis are the standard deviations. }
    \includegraphics[width= 0.98 \columnwidth]{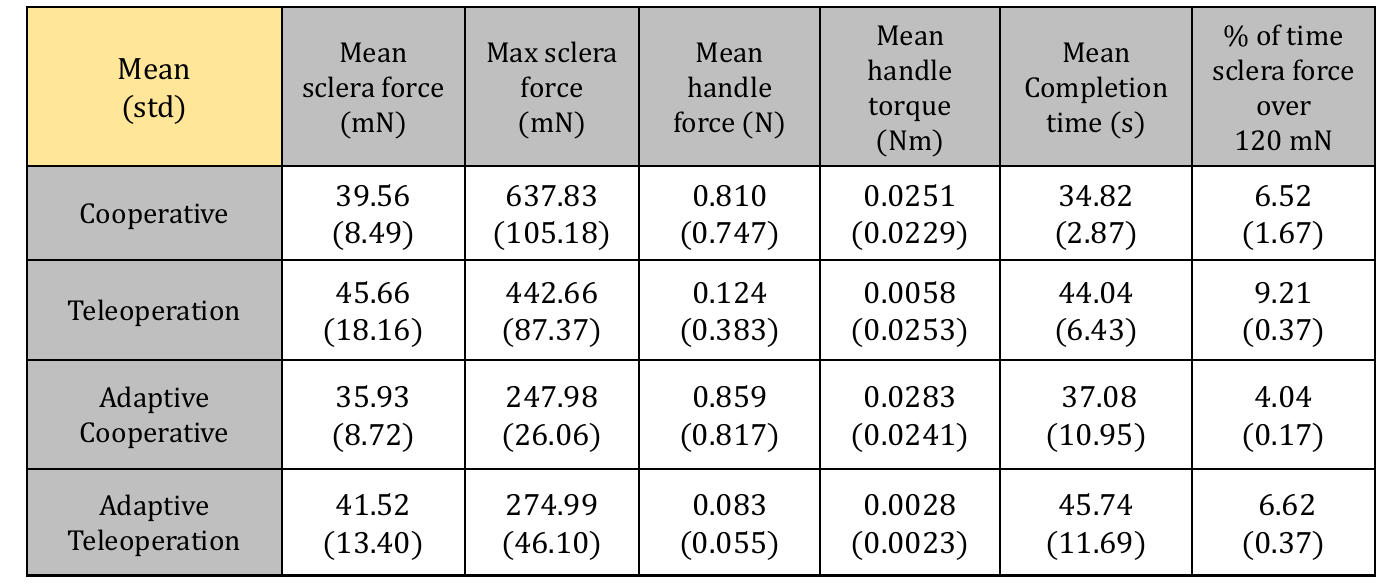}
\label{table:mean_max_force_unilateral_coop_tele}
\end{table}
\vspace{-1pt}
\begin{table}[t!]
\centering
    \caption{P-values of the results of Table \ref{table:mean_max_force_unilateral_coop_tele}. The p-values are calculated using \textit{ttest2} method in MATLAB. Blue: the difference is significant, red: the difference is not significant, black: the difference is not interpretable.}
    \includegraphics[width= 1.00 \columnwidth]{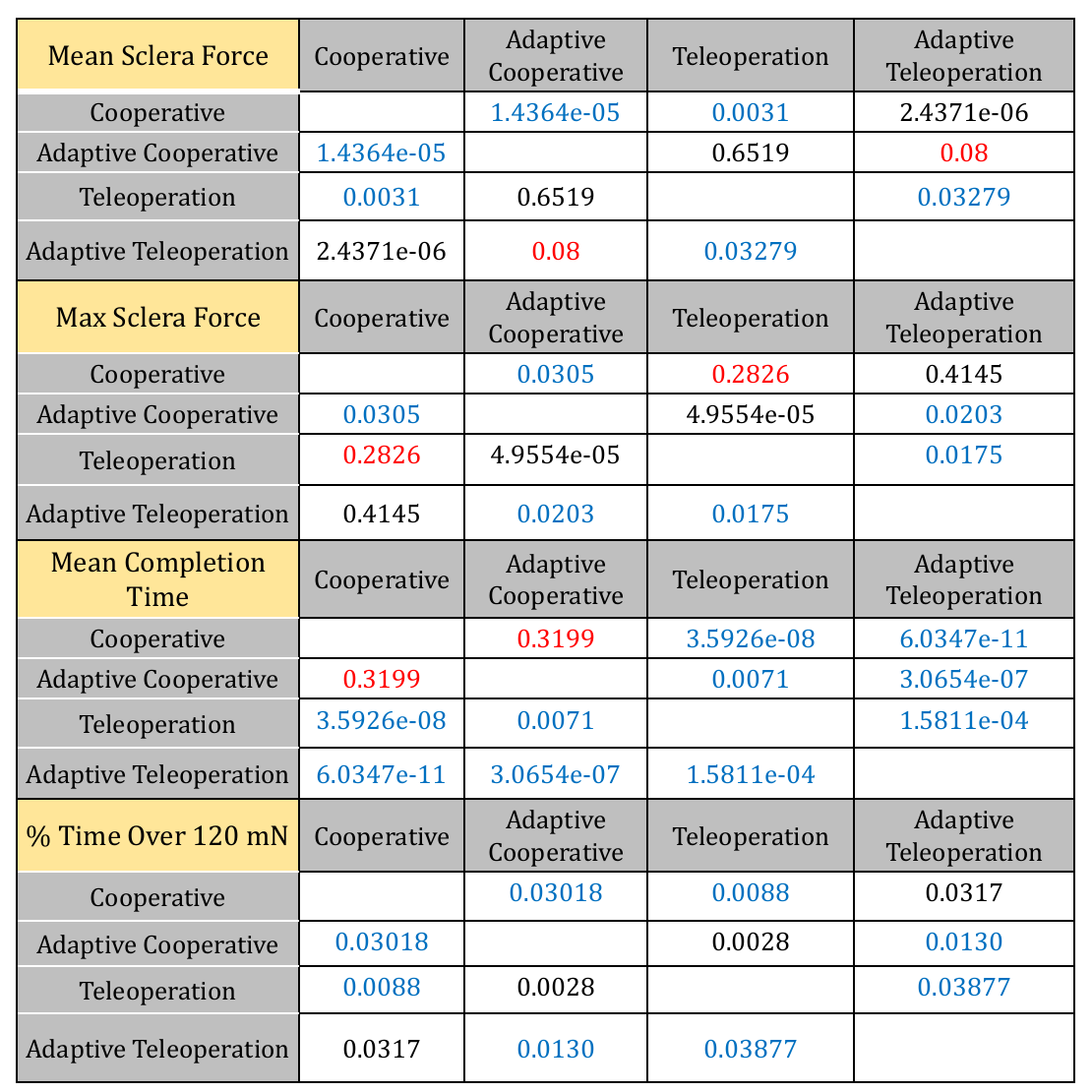}
      \label{table_Pvalues_unilateral_coop_tele}
\end{table}

\begin{figure}
	\centering
    \subfloat[]{
    \centering
    \includegraphics[width=0.48\textwidth]{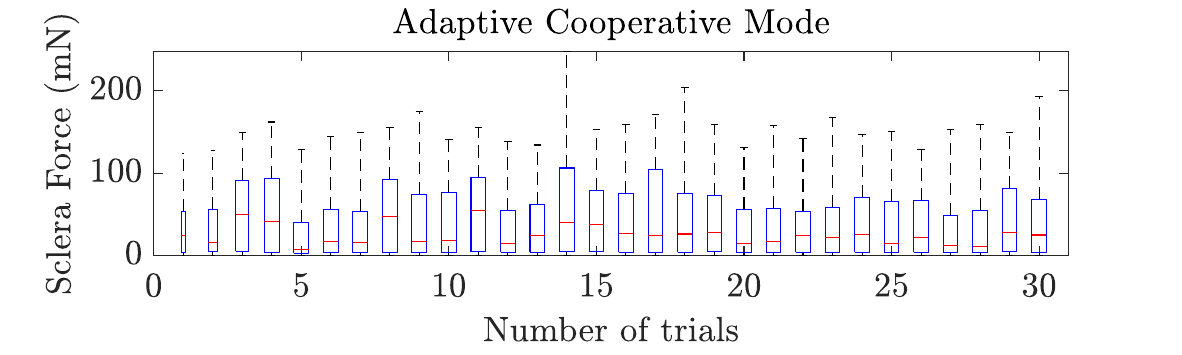}
    \label{fig_scleraForce_cooperative_unilateral}} \\ 
    \subfloat[]{
    \centering
    \includegraphics[width=0.46\textwidth]{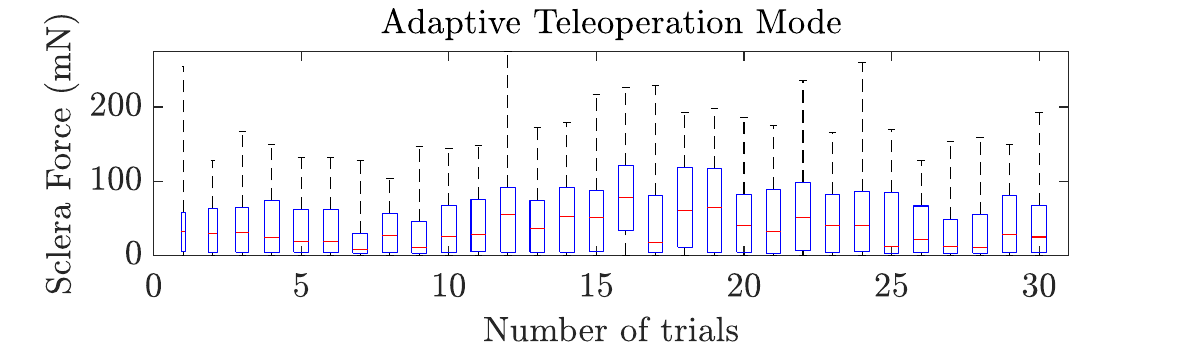}
    \label{fig_scleraForce_teleoperation_unilateral}}
	\caption{Average of the tool-sclera interaction forces for 25 trials for each of the two different control modes: (a) adaptive cooperative and (b) adaptive teleoperation. }
\label{fig_scleraForce_coop_tele_unilateral}
\end{figure}

Table \ref{table:mean_max_force_unilateral_coop_tele} summarizes six key evaluation metrics based on experimental data from 25 trials for each control mode: mean sclera force, maximum sclera force, mean handle force, mean handle torque, mean completion time, and percentage of the aggregate completion times in which the recorded sclera force was above the safety threshold of 120 mN, across all 25 trials. 
Fig. \ref{fig_Sclera_force_adaptive_teleoperation_unilat} shows an example of the sclera force magnitude in one of the adaptive teleoperation trials. Similar behavior is observed in the adaptive cooperative mode. Once each of the sclera force components, $F_{sx}$ and $F_{sy}$, exceeds the adaptive threshold of 100 mN, the adaptive force control algorithm gets activated trying to reduce the sclera force magnitude along a desired exponential trajectory to maintain the sclera force below a safe threshold of 120 mN. This safety feature is missing in the cooperative and teleoperation modes and the sclera force may exceed the 120 mN threshold by a significant amount which could be dangerous for the patient. 
Fig. \ref{fig_scleraForce_coop_tele_unilateral} shows the mean tool-sclera interaction forces for 25 trials for the two adaptive control modes. The mean sclera force for the adaptive cooperative and adaptive teleoperation modes are $35.93\pm 8.72$ mN and $41.52\pm 13.40$ mN, respectively, which are lower than their non-adaptive counterparts. This value for the teleoperation modes is higher than that of the cooperative modes (Table \ref{table:mean_max_force_unilateral_coop_tele}).    
 Also, the maximum sclera force for the adaptive cooperative and adaptive teleoperation modes are $247.98 \pm 26.06$ mN and $274.99 \pm 46.10$ mN, respectively, whereas these values for the cooperative and teleoperation modes are $637.83 \pm 105.18$ mN and $442.66 \pm 87.37$ mN. This demonstrates that the adaptive sclera force control significantly reduced the force applied to the sclera in both cooperative and teleoperation modes. As for the handle force/torque, there is no direct contact between the SHER handle and the user in the teleoperation modes, and the handle force/torque in the teleoperation modes is just the reflection of the sclera force on the handle coordinate which is a very small force compared to that of the cooperative modes (Table \ref{table:mean_max_force_unilateral_coop_tele}). As mentioned, another important metric that shows the surgeon's comfort during interaction with the robot is the mean task completion time. The mean completion time for the teleoperation and adaptive teleoperation modes is $44.04 \pm 6.43$ s and $45.74 \pm 11.69$ s while these values for the cooperative and adaptive cooperative modes are $34.82 \pm 2.87$ s and $37.08 \pm 10.95$ s, respectively, which shows that the cooperative modes are a bit faster than the teleoperation modes. Moreover, for both adaptive cooperative and adaptive teleoperation the percentage of the completion time in which the sclera force is greater than 120 mN is about 30\% less than the cooperative and teleoperation modes, respectively (Table \ref{table:mean_max_force_unilateral_coop_tele}). The corresponding p-values of the results of Table \ref{table:mean_max_force_unilateral_coop_tele} are provided in Table \ref{table_Pvalues_unilateral_coop_tele}. For example, Table \ref{table_Pvalues_unilateral_coop_tele} indicates that the p-value for the mean sclera force between the teleoperation and adaptive teleoperation modes is 0.03279 ($<$0.05) which demonstrates that based on the experimental results the difference between the two control modes is statistically significant. This proves that the adaptive teleoperation mode compared to the teleoperation mode (non-adaptive) can significantly improve patients' safety in terms of mean sclera force being exerted on the eye. However, this difference between the adaptive cooperative and adaptive teleoperation is not statistically significant (since p-value = 0.08 $>$ 0.05), which means that these two modes have similar performance in terms of mean sclera force. Similarly, the difference between the cooperative and adaptive cooperative in terms of the maximum sclera force is statistically significant (p-value = 0.0305) while the difference between the cooperative and teleoperation is not significant (p-value = 0.2826). In terms of the mean completion time, there is no significant difference between the cooperative and adaptive cooperative modes (p-value = 0.3199) whereas, the difference between each cooperative mode and each teleoperation mode is significant (at the p-values $<$ 0.05 level). Of note, these results are collected by a single expert user and cannot be extrapolated to non-expert users (users who are not at their learning curve plateau \cite{zhao2023human}). To the best of our knowledge, the comparison provided above between the cooperative and the teleoperation modes equipped with an adaptive sclera force control algorithm is drawn for the first time in robot-assisted retinal surgery applications.   
\vspace{-1pt}
\section{Conclusion and Future Work} \label{discussion}

In this work, we implemented, for the first time in robot-assisted retinal surgery applications, a novel adaptive sclera force control algorithm in a teleoperation control modality for SHER 2.1 using an FBG-based force-sensing surgical instrument. The performance of this adaptive teleoperation control mode is compared with three previously developed control modes, namely, the cooperative, the teleoperation, and the adaptive cooperative modes in a vessel-following experiment inside an eye phantom. These experiments are conducted unilaterally, and the secondary tool used for rotating the eye phantom has no force-sensing capability. The performance of all four control modes is compared based on several relevant metrics related to the safety of the surgery, and the comfort of the surgeon. These results are reported in Table \ref{table:mean_max_force_unilateral_coop_tele}. They show that both the adaptive cooperative and the adaptive teleoperation modes outperform the cooperative and the teleoperation modes (non-adaptive modes) in terms of the mean and maximum sclera force and the percentage of time at which the sclera force is over 120 mN. The task completion time is shorter for the cooperative modes due to the direct manipulation; however, the teleoperation mode provides additional capabilities such as motion scaling and repositioning that could potentially improve the manipulability and positioning precision.     
For future work, we aim to provide a bilateral teleoperation framework in which two robots, namely SHER 2.0 and SHER 2.1, are both equipped with a force-sensing instrument with adaptive sclera force control capability which improves the safety of the tool-sclera interaction force for the secondary tool as well. Moreover, model predictive control \cite{sadeghnejad2023using}, \cite{amirkhani2020extended, sadeghnejad2023improved} and haptic force feedback may be added to the teleoperation mode such that the sclera force is reflected by the haptic device and perceived by the surgeon.

\vspace{-5pt}
\section{Acknowledgment} 

The authors appreciate Anton Deguet for his technical help in software development.

\bibliographystyle{IEEEtran}
\bibliography{eyebrp}

\end{document}